\documentclass{article} 
\usepackage{iclr2025_conference,times}

\usepackage{amsmath,amsfonts,bm}

\def\eqref#1{equation~\ref{#1}}

\def\1{\bm{1}}

\DeclareMathAlphabet{\mathsfit}{\encodingdefault}{\sfdefault}{m}{sl}
\SetMathAlphabet{\mathsfit}{bold}{\encodingdefault}{\sfdefault}{bx}{n}

\usepackage{hyperref}
\usepackage{hyperref}
\usepackage{amsmath,amssymb,amsfonts}
\usepackage{algorithmic}
\usepackage{graphicx}
\usepackage{textcomp}
\usepackage{xcolor}
\newcommand{\shortname}{LLaVA-Med}
\newcommand{\ourmodel}{R-LLaVA}
\usepackage{times}
\usepackage{latexsym}
\usepackage{pifont}
\newcommand{\xmark}{\ding{55}}
\usepackage[T1]{fontenc}
\usepackage{adjustbox}
\usepackage[utf8]{inputenc}
\usepackage{microtype}
\usepackage{inconsolata}
\usepackage{graphicx}
\usepackage{xcolor}
\usepackage{graphicx}
\usepackage{colortbl}
\usepackage{adjustbox}
\usepackage{enumitem}
\usepackage{comment}
\usepackage{times}
\usepackage{booktabs}
\usepackage{multirow}
\usepackage{latexsym}
\usepackage{rotating} 

\usepackage{tabularx}
\usepackage{subcaption}
\usepackage{colortbl}
\usepackage{amsfonts}
\usepackage{amsmath}
\usepackage[capitalise]{cleveref}
\usepackage[normalem]{ulem}

\title{R-LLaVA: Improving Med-VQA Understanding through Visual Region of Interest}

\author{
\textbf{Xupeng Chen}$^{1}$, \textbf{Zhixin Lai}$^{2}$, \textbf{Kangrui Ruan}$^{3}$, \textbf{Shichu Chen}$^{1}$, \textbf{Jiaxiang Liu}$^{4*}$, \textbf{Zuozhu Liu}$^{4*}$ \\
$^{1}$New York University, $^{2}$Cornell University, $^{3}$Columbia University, $^{4}$Zhejiang University \\
\texttt{\{xc1490, sc10740\}@nyu.edu} \\
\texttt{laizhixin16@gmail.com, kangruir0910app@gmail.com} \\
\texttt{\{jiaxiang.21, zuozhuliu\}@intl.zju.edu.cn}
}

\iclrfinalcopy 
\begin{document}

\maketitle

\begin{abstract}
Artificial intelligence has made significant strides in medical visual question answering (Med-VQA), yet prevalent studies often interpret images holistically, overlooking the visual regions of interest that may contain crucial information, potentially aligning with a doctor's prior knowledge that can be incorporated with minimal annotations (e.g., bounding boxes). To address this gap, this paper introduces \ourmodel{}, designed to enhance biomedical VQA understanding by integrating simple medical annotations as prior knowledge directly into the image space through CLIP. These annotated visual regions of interest are then fed into the LLaVA model during training, aiming to enrich the model's understanding of biomedical queries. Experimental evaluation on four standard Med-VQA datasets demonstrates \ourmodel{}'s superiority over existing state-of-the-art (SoTA) methods. Additionally, to verify the model's capability in visual comprehension, a novel multiple-choice medical visual understanding dataset is introduced, confirming the positive impact of focusing on visual regions of interest in advancing biomedical VQA understanding.
\let\thefootnote\relax\footnotetext{* Corresponding author.}

\end{abstract}

\section{Introduction}

Medical Visual Question Answering (Med-VQA) has recently garnered significant attention \cite{chen2022align, gong2021cross, ren2020cgmvqa, khare2021mmbert}. 
As an emerging area in medical AI, Med-VQA aims to answer medical questions in natural language based on input medical images. A robust Med-VQA system can assist clinicians in interpreting medical images, thereby ensuring accuracy and expediting the diagnostic process. For patients, automated Med-VQA services can greatly meet the demand for personalized healthcare consultations \cite{liu2023parameter}.

Numerous deep learning-based approaches have been explored in the realm of Med-VQA \cite{tiong2022plug, banerjee2020weaqa, changpinyo2022all, liu2023chatgpt, gai2024medthink,liu-etal-2024-medcot}. For instance, \citet{nguyen2019overcoming} utilized Bilinear Attention Networks (BAN) \cite{kim2018bilinear}, enhancing them with Mixed Enhanced Visual Features (MEVF), which integrates pre-trained meta-learning modules and Convolutional Denoising Autoencoders (CDAE) to improve the performance of Med-VQA models. Building on this, \citet{zhan2020medical} proposed a conditional reasoning framework to further enhance the inference capabilities of Med-VQA models. However, many of these methods tend to perform poorly in practical scenarios, mainly due to limitations in the extraction and integration of information from a limited number of medical images and text data \cite{eslami2021does, song2022clip, wang2022clip,liu2024vpl,liu2025kpl}. To address this, \citet{eslami2021does} introduced the CLIP architecture into the MEVF framework \cite{nguyen2019overcoming}, using CLIP as the visual encoder pre-trained in the ROCO multimodal medical dataset \cite{pelka2018radiology}, which demonstrated significant performance gains. Additionally, \citet{liu2023parameter} developed VQA-Adapter, a lightweight adapter that, combined with label smoothing, efficiently fine-tunes CLIP for Med-VQA, reducing computational costs and mitigating overfitting. \citet{li2024llava} proposed LLaVA-Med, leveraging GPT-4 and a novel curriculum learning approach to efficiently train LLaVA in biomedical images, further enhancing the capabilities of Med-VQA.

\begin{figure}[t]
\centering
  \includegraphics[width = 0.5 \linewidth]{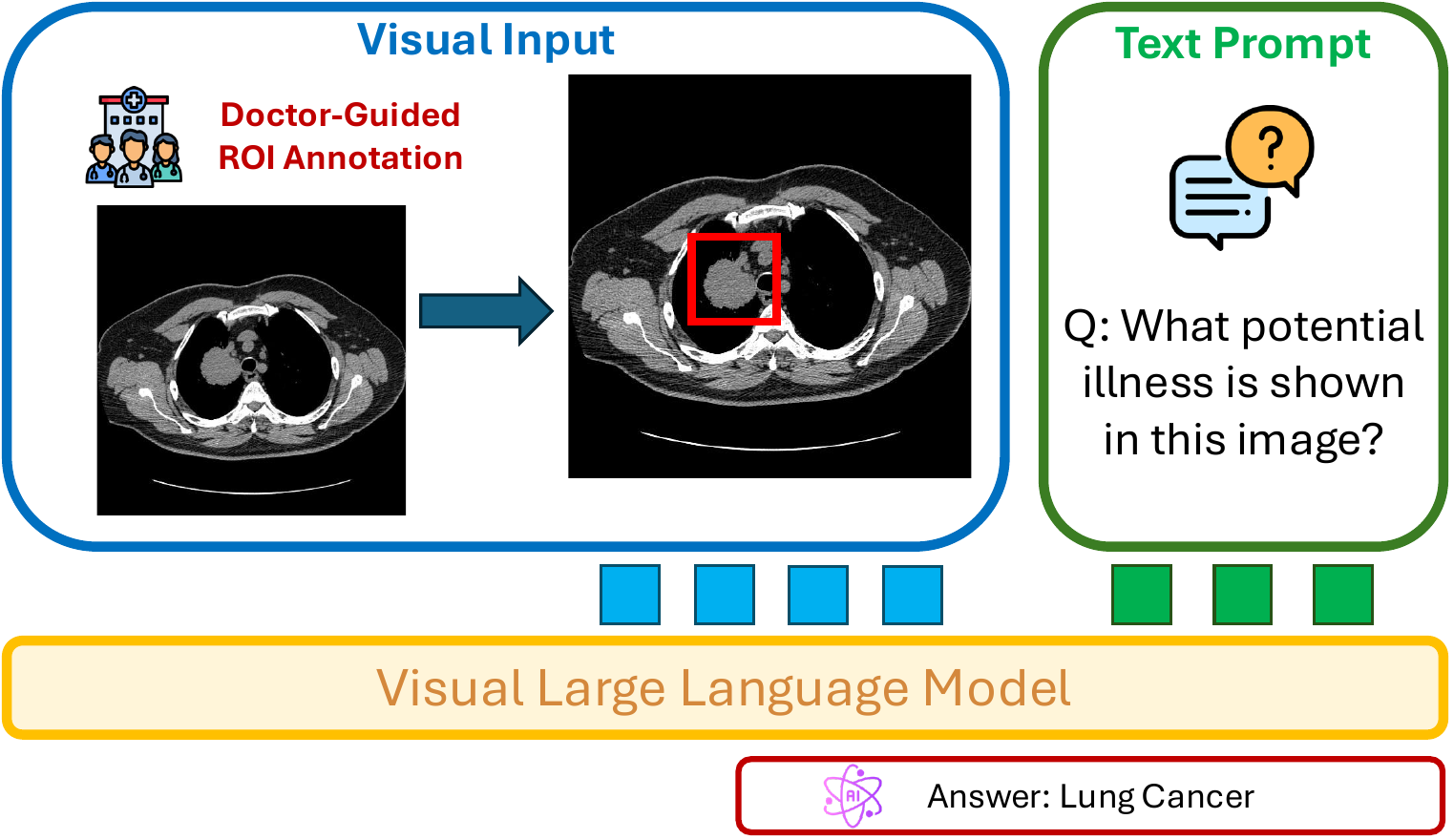}
  \caption{Integration of doctor-guided Region-of-Interest (RoI) and original image for multimodal conversational assistance. Specifically, a Doctor-Guided ROI, which is highlighted in the red box, is overlaid onto the original image in line with the clinician's diagnostic approach.}
 \label{fig:main_idea}
\end{figure}

However, previous Med-VQA approaches have often overlooked the importance of medical and spatial priors that are essential for disease localization in Med-VQA tasks. Inspired by clinical practices in which doctors rely on such priors, we propose the \ourmodel{} framework, which incorporates doctors' domain knowledge and spatial annotations into Med-VQA models. \cref{fig:main_idea} shows the main idea of the proposed method.
Specifically, we integrate simple physician annotations, such as bounding boxes (bbox), as prior knowledge, directly injecting these visual cues into the image space via CLIP 
showing CLIP's ability to interpret visual markers. These annotated regions are then fed into the LLaVA model during training. Our experiments show that even minimal annotations from doctors can significantly improve the accuracy of the model.
\ourmodel{} consistently outperforms the state-of-the-art Med-VQA methods in four large-scale datasets, demonstrating remarkable performance improvements.
Our work has two major contributions:

\begin{itemize}
    \item The work introduces \ourmodel{} method to improve Med-VQA by incorporating regions of visual interest through a two-phase training process.

    \item To validate the model's visual comprehension, We introduce a multiple-choice medical visual understanding dataset, confirming the positive impact of integrating Visual Regions of Interest on enhancing Biomedical VQA Understanding.
\end{itemize}

\section{Method}

R-LLaVA is based on the premise that visual LLMs should analyze not only the visual content of the images themselves but also the specific regions of interest highlighted by clinicians.
In this section, we detail our approach, starting with the reconstruction of medical VQA datasets to incorporate region-based information, simulating how clinicians annotate critical regions of interest. 
Following this strategy of dataset reconstruction, we explain the process utilized to train \ourmodel{} based on these annotations.

\subsection{Medical Dataset Reconstruction with Region-of-Interest (RoI) VQA}\label{method:data_generation}

In Med-VQA, utilizing datasets incorporating region-based information is helpful for enabling models to accurately focus on and interpret specific regions of medical images. 
This targeted focus ensures that model responses are grounded in the precise anatomical or pathological features of interest, leading to more accurate and clinically meaningful answers in complex medical scenarios.
To evaluate and enhance the region-learning capabilities of Vision-and-Large-Language Models(VLLMs) in medical VQA tasks, we propose a strategy to reconstruct the existing medical VQA datasets by integrating RoI VQA pairs. These pairs are designed to improve and evaluate the model’s ability to localize and describe specific regions within medical images. 

\begin{figure*}[!t]
\centering
\includegraphics[width=\textwidth]{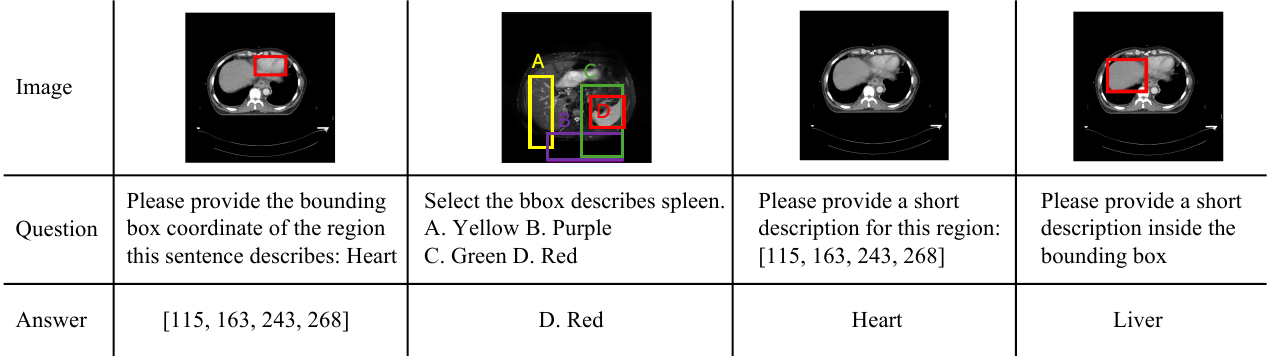}
  \caption{Region-of-Interest (RoI) QA from Reconstructed VQA Dataset}
\label{fig:data_reconstruction}
\vspace{-1em}
\end{figure*}

As is shown in~\cref{fig:data_reconstruction}, the reconstructed dataset comprises four types of QA pairs: (1) Region localization, where the model is required to predict the bounding box coordinates corresponding to a described region, e.g.,  "Please provide the bounding box coordinate of the region this sentence describes: Heart". 
(2) Region selection - where the model is required to select the bounding box among four corresponding to a described region, e.g., "Select the bounding box (bbox) describes spleen. A. Yellow B. Purple C. Green D. Red". (3) Region description with bounding box coordinates - where the model is asked to provide a description given a bounding box, e.g., "Please provide a short description for this region: [115, 163, 243, 268]". (4) Region description with Bounding Box highlighted in input image, where the model is asked to describe the bounding box, e.g., "Please provide a short description inside the bounding box".

This reconstruction of Med-VQA datasets with RoI annotations aims to push the boundaries of current VLLMs, providing a more rigorous evaluation framework for their region-learning capabilities in medical imaging tasks.

\subsection{\ourmodel{}  Training Stage I: Pre-training} 

The first stage of training is to align the biomedical concepts while maintaining efficiency. As shown in~\cref{fig:training_pipeline}(a), the image encoder (e.g., CLIP \citep{radford2021learning}) and the language model (e.g. Vicuna \citep{zheng2023judging}) are frozen in stage I. Only the multi-modal connector is trained by updating the projection matrix. This freezing technique is crucial for efficiency and understanding concepts, as it allows the model to connect the foundational representations of visual and textual embeddings under medical settings. 
The major dataset used at this stage follows the LLaVA-Med alignment dataset~\cite{li2023llavamedtraininglargelanguageandvision}. It is built upon a large-scale dataset of 15 million parallel figure-caption pairs for biomedical vision-language processing—PMC-15M. PMC-15M is shrunk to only 600,000 pairs to balance concept coverage and efficiency, which speeds up training while still providing great alignment performance. 

\subsection{\ourmodel{} Training Stage II: Instruction Tuning with Visual RoI Approach} 

In Stage II, the model is aimed to learn to follow various types of textual instructions across a wide range of specific medical fields and complete field-specific tasks. The model is also designed to focus on the Region-of-Interest (RoI) (e.g., bounding boxes) to enhance its vision ability for Med-VQA. 

\begin{figure*}[!t]
\centering
  \includegraphics[width=\textwidth]{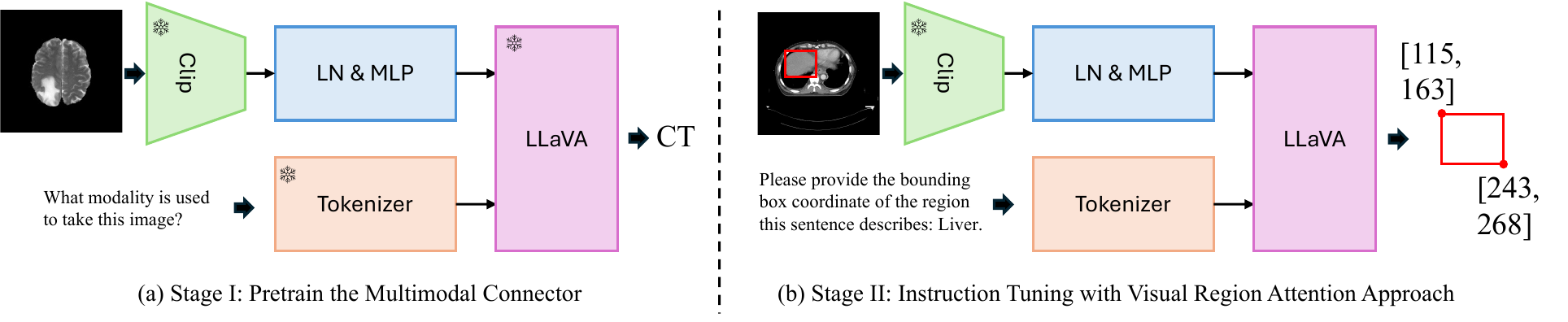}
  \caption{R-LLaVA Training Pipeline (Stage I and Stage II)}
  \vspace{-2mm}
  \label{fig:training_pipeline}
\end{figure*}

As illustrated in~\cref{fig:training_pipeline}(b), during stage II, the visual encoder is frozen while the pre-trained weights of the projection layer and the language model are updated. 
To enhance the model's capacity for instruction-following and task execution in a biomedical conversational context, we further fine-tune the model on biomedical language-image instruction-following data, improving its ability to handle task transfer across established Med-VQA datasets.

For certain biomedical applications, it is essential to create highly accurate, dataset-specific models to meet the required performance standards.
After completing the pretraining based on LLaVA-Med, we trained on four distinct Med-VQA datasets, each covering a range of dataset sizes and specialized medical topics.
Given a biomedical image as input, the model is tasked with answering multiple natural language questions, generating responses in free-form text for both close-set and open-set question types.

To strengthen the model’s ability to interpret arbitrary visual regions of interest, we utilize CLIP’s built-in capacity to encode both the image and supplementary human-annotated visual markers. This provides enhanced guidance by merging potential doctor annotations into the image via alpha blending, drawing focus to regions of interest:
\[
\bar{\mathbf{X}}_{\mathrm{merged}} = \alpha \cdot \mathbf{P}_{\mathrm{doc}} + (1 - \alpha) \cdot \mathbf{X}_{\mathrm{img}}
\]
where \(\alpha\) controls the transparency level, \(\mathbf{X}_{\mathrm{img}}\) is the base image, and \(\mathbf{P}_{\mathrm{doc}}\) represents the medical regions of interest. The composite image \(\bar{\mathbf{X}}_{\mathrm{merged}}\) is then fed into the multimodal language model to guide its attention toward these key areas \cite{vaswani2017attention,dosovitskiy2020vit,ruan2024infostgcan}.

To capture both detailed and abstract information, we extract multi-level features from several layers of CLIP. The shallower layer is used for capturing finer details, while deeper layers capture more abstract semantic representations \citep{ruan2024s2e}. These features are combined, normalized using Layer Normalization for stability, and then processed by a Multi-Layer Perceptron(MLP) layer to integrate the diverse visual cues effectively.

This direct integration of visual prompts over regions of interest brings several benefits. It reduces model complexity by eliminating additional processing components and mirrors natural doctor-patient interactions, making it suitable for real-world applications.

We employ autoregressive language modeling to train the model, optimizing the likelihood of producing the ground-truth answer tokens \(\mathbf{Y}\):
\begin{align}
\begin{split}
    & P(\mathbf{Y} \mid \bar{\mathbf{X}}_{\mathrm{merged}}, \mathbf{X}_{\text{instruct}}) \\
    & = \prod_{t=1}^{T} P_{\boldsymbol{\Omega}}(x_t \mid \bar{\mathbf{X}}_{\mathrm{merged}}, \mathbf{X}_{\text{instruct}}, \mathbf{Y}_{<t})
\end{split}
\end{align}
where \(\bar{\mathbf{X}}_{\mathrm{merged}}\) contains the integrated images with an emphasis on regions of interest, \(\mathbf{X}_{\text{instruct}}\) provides the textual instruction, which can be constructed from questions. \(T\) is the sequence length of the target answer \(\mathbf{Y}\), \(\mathbf{Y}_{<t}\) includes all the answer tokens preceding the current prediction token \(x_t\), and $\boldsymbol{\Omega}$ represents the trainable parameters.

This training approach equips the model to generate accurate and context-aware responses by integrating visual content, instructions, and visual regions of interest directed toward specific tissues, organs, and so on. 
This process imitates the interactions between doctors and patients by highlighting diagnostically relevant areas. 
It is especially effective for tasks that require a nuanced understanding of both the visual elements within a specific tissue or organ and the user’s intent conveyed through certain visual markers.


\section{Experiment}

In this section, we first provide a detailed description of the utilized dataset (\Cref{subsec:dataset}), including the types of questions it encompasses. 
Next, we present the selected evaluation metrics (\Cref{subsec:metrics}). Based on the chosen datasets and metrics, we outline the training strategies and settings employed in our experiments (\cref{subsec:strategy}). 
We present the main results in \Cref{subsec:main_results}, along with a comprehensive ablation study analyzing the impact of our methodological choices (\Cref{subsec:ablation_study}). 

\subsection{Dataset} \label{subsec:dataset}

For pretraining, we utilize a large-scale medical image-caption dataset from \cite{li2024llava}, containing 600K pairs to facilitate foundational alignment with medical concepts. For Fine-tuning, we employ four VQA datasets: VQA-RAD, SLAKE-EN, PathVQA, and VQA-Med. We augment the SLAKE-EN dataset with RoI QA, as detailed in \cref{method:data_generation}.
\noindent  We adopt a consistent 80/20 train-test split for all datasets, including VQA-RAD, SLAKE-EN, PathVQA, and VQA-Med 2019.

We define three question types: close-ended, multi-choice, and open-ended. 
Close-ended questions yield straightforward answers, typically "Yes" or "No" (e.g., "Is there any abnormality in the spleen?"). Multi-choice questions provide a set of predefined answers (e.g., "Which rectangle contains the object representing Cardiomegaly?"). 
Open-ended questions solicit descriptive responses (e.g. "Please provide a short description for this region").

\begin{table*}[ht!]
\centering
\resizebox{\textwidth}{!}{
\begin{tabular}{ll|cc|ccc|cc|c|c}  
 & & \multicolumn{2}{c|}{\bf VQA-RAD} & \multicolumn{3}{c|}{\bf SLAKE-EN} & \multicolumn{2}{c|}{\bf PathVQA} & \multicolumn{1}{c|}{\bf VQA-Med 2019} &\multicolumn{1}{c}{\bf Act.} \\
Method  &  & Open   & Closed    & Open   & Closed & Multi & Open   & Closed  &  Closed & \\
\hline
\multicolumn{9}{l}{\it Performance of prior methods with reported metrics in the study} \\  
\hline
{CGMVQA Ens.} \cite{ren2020cgmvqa}  &    & -       & -    & - & - & -  & - & -   & \underline{78.10}          & -     \\
{MMBERT} \cite{khare2021mmbert}    &   & -          & 77.90   & -      & -      & -             & -        & -         & \underline{78.10}          & -               \\

M2I2~\citep{li2022self} &   & - & 83.50 & - & 91.10 &- &- & 88.00 &    - &- \\
CLIP-ViT w/ GPT2-XL \cite{van2023open} & & - & - & 84.30 & 82.10 & -&40.0 & 87.00 &    - & - \\
VL Encoder–Decoder~\citep{bazi2023vision} & & - & 82.47 & - & - & - & - & 85.61 &    - & - \\
Q2ATransformer~\citep{liu2023q2atransformer} & & - & 81.20 & - & - &- &54.85 & 88.85 &    - & - \\
Prefix T. Medical LM~\citep{van2023open} & & - & - & - & 82.01 &- &- & 87.00 &     - & - \\
PubMedCLIP~\citep{eslami2023pubmedclip} & & - & 80.00 & - & 82.50 & -&- & - &   - &   - \\
BiomedCLIP~\citep{zhang2023large} & & - & 79.80 & - & 89.70 & -&- & - &    - & - \\

BiomedGPT-S~\citep{zhang2023biomedgpt} &  & 13.40 & 57.80  & 66.50 & 73.30 &- &10.70 & 84.20 &    - & - \\
BiomedGPT-M~\citep{zhang2023biomedgpt}&  & 53.60 & 65.07 & 78.30 & 86.80 &- & 12.5 & 85.70 &    - & - \\
Gemini Pro \cite{yang2024advancing}  &     & -         & 60.29    & -        & 72.60     & -        & -  & 60.22             & 70.30      & -        \\
\hline
\multicolumn{9}{l}{\it Supervised finetuning results} \\
\hline
LLaVA \cite{liu2024visual}& & 50.00 & 65.07 & 78.18 & 63.22 & -&7.74 & 63.20  &     - & 7B \\
\shortname{} (LLama7B) \cite{li2024llava} & &  61.52 & \textbf{84.19} & 83.08 & \underline{85.34}   &- &   37.95    & {91.21} &    - & 7B \\
\shortname{} (Vicuna7B)  \cite{li2024llava}&  &  \underline{64.39} & 81.98 & \underline{84.71}  & 83.17  &  \underline{40.85} & \bf 38.87  & \underline{91.65} &    - & 7B \\
\shortname{} (Phi2.7B) \cite{li2024llava} &   & 54.83 & {81.35} & 81.29  &  83.29  &- & {31.73} & 90.17  &   - & 2.7B \\

\rowcolor[HTML]{CBC5D3} \ourmodel{} (7B)     & &  \textbf{64.91}	&\underline{83.76}  & \textbf{89.47}	& \textbf{90.13}& 	\textbf{68.85}  & \underline{38.24}	 & \textbf{92.59}   &    \textbf{80.97}  & 7B  \\

\hline
\end{tabular}
}
\caption{Performance on Med-VQA tasks. \textbf{Bold} denotes the best performance; \underline{underlined} denotes the second-best.}
\label{tab:main_table}
\vspace{-1em}
\end{table*}

\subsection{Metrics} \label{subsec:metrics}

For close-ended and multi-choice questions, we use accuracy as the evaluation metric, assessing it by directly comparing predictions with ground truth. For open-ended questions, we measure recall, calculating the proportion of ground-truth tokens present in the generated sequences \citep{chen2024llava,jiang2024mixtral}. 

\subsection{Training Strategy and Setting} \label{subsec:strategy}

We use CLIP-ViT-L/14@336px \cite{radford2021learning} as the vision encoder to extract relevant features from input medical images, and Vicuna 1.5 \cite{chiang2023vicuna} as the language model. A 2-layer MLP serves as the multi-modal projector.

\noindent \textbf{Stage I: Pretraining.} The 2-layer multimodal projector is pre-trained on large-scale medical image-caption pairs, developing a foundational understanding of visual data without instruction-following abilities. The training uses a batch size of 256, a learning rate of $1 \times 10^{-3}$, and a maximum sequence length of 2048 tokens for 1 epoch without weight decay.

\noindent \textbf{Stage II: Instruction Fine-Tuning.} With the CLIP encoder fixed, we fine-tune the remaining model components on specialized biomedical instruction-following data (up to 60K samples) with diverse queries and inline mentions, enhancing performance across medical domains. This stage uses a batch size of 128, a reduced learning rate of $2 \times 10^{-5}$, and a maximum sequence length of 2048 tokens, trained for 1 epoch without weight decay.

Both pretraining and fine-tuning stages are conducted on 8 A100 80G GPUs. Pretraining requires approximately 6 hours for the 7B model and 12 hours for the 13B model, while fine-tuning takes 3 hours for the 7B model and 6 hours for the 13B model. All models use FP16 precision for both training and inference.


\subsection{Main Results} \label{subsec:main_results}

\paragraph{Quantitative Comparison} The experimental results in \cref{tab:main_table} demonstrate that our proposed model, \ourmodel(7B), sets a new state-of-the-art on several medical visual question answering (VQA) benchmarks. Specifically, \ourmodel(7B) achieves the best performance on SLAKE-EN, with an accuracy of 89.47\% on open-ended questions and 90.13\% on close-ended questions, outperforming all competing models by a significant margin. This highlights the effectiveness of our region of interest approach. Furthermore, \ourmodel(7B) also delivers superior results on VQA-Med 2019, achieving an accuracy of 80.97\% on close-ended questions. For the VQA-RAD and PathVQA datasets, our model consistently achieves superiority on both open-ended and closed-ended tasks, reinforcing its robustness across diverse medical VQA challenges.

\begin{figure}[!ht]
\centering
  \includegraphics[width = 0.6 \linewidth]{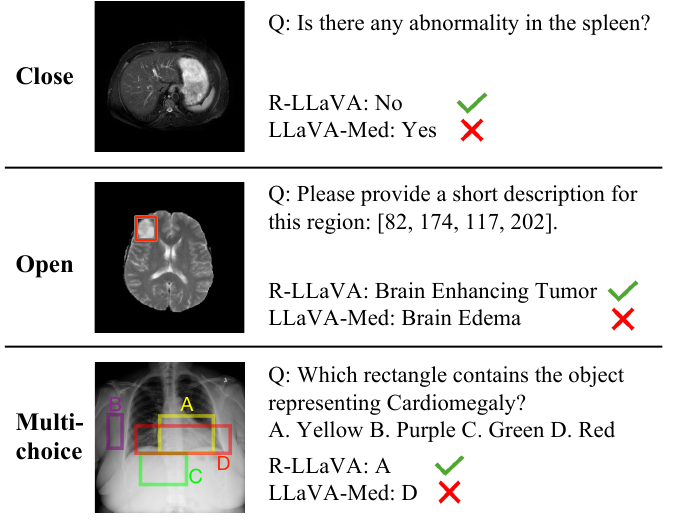}
  \caption{Qualitative comparison of medical visual question answering on three types of questions from SLAKE dataset.}
 \label{fig:result_example}
\end{figure}

\begin{table*}[ht!]
\centering
\resizebox{0.95\textwidth}{!}{ 
\begin{tabular}{lll|cc|ccc|cc|c}  
 & & & \multicolumn{2}{c|}{\bf VQA-RAD} & \multicolumn{3}{c|}{\bf SLAKE-EN} & \multicolumn{2}{c|}{\bf PathVQA} & \multicolumn{1}{c}{\bf Med 2019}   \\
Data & Init  & Size  & Open   & Closed    & Open   & Closed & Multi & Open   & Closed  &  Closed  \\  
\hline
single & ViP-LLaVA & 7B & 62.78 & 80.27 & 88.61 & 88.94 & \bf 70.65 & 35.63 & 91.02 & 80.35 \\
single & ViP-LLaVA & 13B & 60.58 & 77.19 & 87.07 & 85.54 & 66.57 & 36.4 & 90.35 & 77.51 \\
single & Vicuna & 7B  & 62.26 & 78.17 & 87.55 & 88.55 & 68.34 & 32.97 & 90.93 & 79.38 \\
single & Vicuna & 13B & 58.3 & 76.13 & 85.99 & 86.58 & 64.47 & 33.92 & 89.17 & 76.2 \\
all & ViP-LLaVA & 7B  &\textbf{64.91} &  83.76  & \textbf{89.47} & \textbf{90.13} &  68.85  & \textbf{38.24} & \textbf{92.59} & \textbf{80.97} \\
all & ViP-LLaVA & 13B & 63.49 & \bf 84.31 & 87.14 & 87.31 & 69.14 & 36.29 & 90.7 & 79.14 \\
all & Vicuna & 7B  & 59.99 & 80.64 & 87.28 & 89.48 & 67.25 & 37.21 & 91.51 & 77.69 \\
all & Vicuna & 13B & 57.15 & 81.03 & 86.31 & 88.44 & 65.33 & 35.22 & 89.32 & 77.79 \\
\hline
\end{tabular}
}
\caption{Ablation Studies on Model Choices}
\label{tab:ablation_models}
\end{table*}

\begin{table*}[ht!]
\centering
\resizebox{0.95\textwidth}{!}{ 
\begin{tabular}{cc|cc|ccc|cc|c}  
 & &  \multicolumn{2}{c|}{\bf VQA-RAD} & \multicolumn{3}{c|}{\bf SLAKE-EN} & \multicolumn{2}{c|}{\bf PathVQA} & \multicolumn{1}{c}{\bf Med 2019}   \\
Pretrain & Finetune  & Open   & Closed    & Open   & Closed & Multi & Open   & Closed  &  Closed  \\  
\hline
\xmark  & \xmark  & 20.74&  59.19& 26.82 &50.24& - & 8.74& 45.65 & - \\
\checkmark  & \xmark & 27.88 & 57.81 & 40.05 & 45.22 & 27.49 & 14.25 & 50.53 & 42.38    \\
\xmark  & \checkmark & 50.25 & 64.53 & 70.27 & 73.97 & 50.73 & 22.84 & 60.49 & 58.67  \\
\checkmark  & \checkmark  & \textbf{64.91} & \textbf{83.76} & \textbf{89.47} & \textbf{90.13} & \textbf{68.85} & \textbf{38.24} & \textbf{92.59} & \textbf{80.97} \\
\hline
\end{tabular}
}
\caption{Ablation Studies on Training Strategies}
\label{tab:ablation_table2}
\vspace{-1em}
\end{table*}

\paragraph{Qualitative Comparison} From \cref{fig:result_example}, \ourmodel{} demonstrates superior performance over LLaVA-Med across all question types. Specifically, for close-ended questions, such as identifying abnormalities in specific organs (e.g., spleen), \ourmodel{} provides more accurate responses. For open-ended questions that require descriptive answers, \ourmodel{} accurately describes specific regions with detailed terms closely aligned to medical findings. For example, it correctly identifies a "Brain Enhancing Tumor," demonstrating a deeper understanding of the medical context, whereas LLaVA-Med misclassifies it as "Brain Edema," highlighting a gap in spatial and contextual comprehension. In multi-choice scenarios, \ourmodel{} consistently selects the correct bounding box corresponding to medical conditions like cardiomegaly. Its precision in choosing the right option reflects its effective use of attention-based mechanisms to interpret complex image regions, surpassing LLaVA-Med’s less consistent performance. 
These quantitative and qualitative results underscore \ourmodel{}’s enhanced ability to handle the nuanced demands of Med-VQA tasks.

\subsection{Ablation Study} \label{subsec:ablation_study}

In this section, we evaluate the effectiveness of each component in the model training process, covering model selection, two-stage training strategy, and the visual Region-of-Interest (RoI) approach.

\subsubsection{Model Selection}

We first conducted an ablation study on model configurations, focusing on whether the initial parameters were loaded from \textbf{Vicuna} \citep{zheng2023judging} or \textbf{ViP-LLaVA} \citep{cai2024vip}. 
Additionally, we compared two fine-tuning strategies: \textit{single} and \textit{all}. 
In the \textit{single} strategy, the model is trained and evaluated independently on each dataset. In contrast, the \textit{all} strategy involves training the model on all datasets collectively, enabling it to generalize and better understand medical prompts across different datasets. Furthermore, we evaluated two model sizes: \textbf{7B} and \textbf{13B}. 

As shown in \cref{tab:ablation_models}, the 7B model, initialized with parameters from ViP-LLaVA and fine-tuned on data from all four datasets, achieves the best performance.

\begin{table*}[ht!]
\centering
\resizebox{0.95\textwidth}{!}{ 
\begin{tabular}{cc|cc|ccc|cc|c}  
 & & \multicolumn{2}{c|}{\bf VQA-RAD} & \multicolumn{3}{c|}{\bf SLAKE-EN} & \multicolumn{2}{c|}{\bf PathVQA} & \multicolumn{1}{c}{\bf Med 2019}   \\
Bbox in Prompt & alpha  & Open   & Closed    & Open   & Closed & Multi & Open   & Closed  &  Closed  \\  
\hline
\xmark &  $\alpha = 0$  & 63.26 & 82.22 & 80.39 & 81.32 & 31.06 & 34.47 & 90.69 & 77.13   \\
\checkmark &   $\alpha = 0$  & 63.05 & 82.11 & 83.41 & 81.86 & 25.22 & 35.78 & 89.68 & 78.91   \\
\checkmark & $\alpha = 96$ & 64.73 & 83.09 & 88.65 & 89.32 & 65.32 & 36.84 & 90.95 & 80.12 \\
\checkmark & $\alpha = 128$ & 64.49 & 82.57 & 88.88 & 88.98 & 67.41 & 37.11 & 90.47 & 80.31   \\
\checkmark & $\alpha = 255$ & 64.79 & 82.72 & 89.69 & 89.08 & 67.83 & 37.50 & 91.84 & 80.79    \\
\checkmark & Dynamic  &   \textbf{64.91} & \textbf{83.76} & \textbf{89.47} & \textbf{90.13} & \textbf{68.85} & \textbf{38.24} & \textbf{92.59} & \textbf{80.97} \\
\hline
\end{tabular}
}
\caption{Ablation Studies on Region-of-Interest (RoI) Approach}
\label{tab:ablation_table3}
\end{table*}

\subsubsection{Two-stage Training Strategy}

We demonstrate the importance of both training stages: pretraining on the LLaVA-Med dataset and fine-tuning across four VQA datasets.

From~\cref{tab:ablation_table2}, we observe that without pretraining and fine-tuning (row 1), the model achieves very low accuracy across all four datasets, with open-ended question accuracy below 30\% on the first three datasets and only 8.74\% on PathVQA. With pretraining only (row 2) or fine-tuning only (row 3), the performance improves. Fine-tuning alone generally yields better results across all datasets compared to pretraining alone; for instance, it achieves an accuracy of 70.27\% on SLAKE-EN open-ended questions versus 40.05\% for pretraining only and 26.82\% without any training. However, pretraining only slightly reduces accuracy on close-ended questions (57.81\% for VQA-RAD and 45.22\% for SLAKE-EN) compared to the baseline without training (59.19\% for VQA-RAD and 50.24\% for SLAKE-EN). With both training stages (row 4), our model achieves optimal results, showing double-digit improvements across various question types and datasets compared to other experiments.


Our findings highlight the importance of performing both pretraining and fine-tuning. Fine-tuning enables the model to acquire task-specific abilities, while pretraining provides essential background knowledge on domain-specific topics, which significantly enhances the performance compared to fine-tuning alone.

\subsubsection{Visual Region-of-Interest (RoI) Approach}

We demonstrate the effectiveness of the visual RoI approach through both quantitative results in \cref{tab:ablation_table3} and qualitative results in \cref{fig:result_example}. In \cref{tab:ablation_table3}, "Bbox in Prompt" means adding bounding box information in the prompt, and "alpha" is the weight of alpha blending. 

\paragraph{Visual Region-of-Interest (RoI) Approach} From \cref{tab:ablation_table3}, without bounding box information in both prompt and input image (Row 1), we observe an accuracy of 80.39\% on open-ended questions, 81.32\% on close-ended questions, and 31.06\% on multi-choice questions for the SLAKE-EN dataset. The lack of bounding box information in both the prompt and input image results in lower performance across all datasets, particularly in the multi-choice questions, than in the experiments with Visual RoI (Row 3, 4, 5, 6). Adding bounding box information in the prompt while keeping alpha at 0 based on Row 1, Row 2 slightly improves performance across open-ended questions (83.41\%) and close-ended questions (81.86\%) in SLAKE-EN, with a minor boost in multi-choice questions (25.22\%). This indicates that bounding box prompts contribute positively but are insufficient alone. 

These findings demonstrate that the visual RoI approach, which incorporates bounding boxes in both the prompt and input image, is critical for enhancing model performance on VQA datasets, particularly in the multi-choice categories of SLAKE-EN. 

\textbf{Alpha Blending Strategy} This experiment analyzes the impact of varying bounding box opacity (alpha) on VQA performance across multiple datasets. Higher alpha values represent a more visible bounding box, while lower values signify a subtler presence. Alpha values include fixed levels (96, 128, 255) and a randomly sampled range within [96, 255]. For all VQA datasets, alpha values of 96, 128, and 255 yield similar results, with slight incremental improvements as alpha increases. This suggests that a more visible bounding box can aid in certain tasks but may not significantly influence outcomes. Randomized Alpha (Dynamic) within the range [96, 255] achieves the best results across all VQA datasets compared to other fixed alpha groups. The variability introduced by random alpha values likely provides a more robust learning experience by exposing the model to diverse bounding box appearances, resulting in improved generalization and accuracy.

\section{Related Work}

\subsection{Visual Large Language Model} 

Recent years have witnessed significant advancements in Large Language Models (LLMs) \cite{brown2020language}. Concurrently, developments in Large Vision Models (LVMs) have emerged, complementing LLMs through their capacity for visual understanding.
Integrating multimodal understanding into large language models has significantly advanced multimodal AI capabilities \cite{ruan2024twitter,ruan2025causal}. 
Early models like ViLBERT \cite{lu2019vilbert} and VisualBERT \cite{li2019visualbert} extended the BERT architecture to process both text and images, enabling tasks such as visual question answering and image captioning. 
Contrastive learning frameworks like CLIP \cite{radford2021learning} learned joint representations of images and text from large-scale datasets, achieving impressive zero-shot performance on various tasks. 
Recent developments focus on a more seamless integration of vision and language. Models like Flamingo \cite{alayrac2022flamingo} and PaLI \cite{chen2022pali} incorporate visual information into language models using gated cross-attention and scale-up pre-training with multilingual data. 
Furthermore, the enhanced multimodal features of ChatGPT and Gemini \cite{team2023gemini} model represent significant strides toward more integrated and capable AI systems.

\subsection{Med-VQA}

Med-VQA is a method where a model answers questions from patients or clinicians based on medical images like CT scans, MRI scans, or pathology images.
With the advancement of deep learning, researchers have proposed numerous Med-VQA methods.
M2I2 ~\cite{li2022selfsupervisedvisionlanguagepretrainingmedical} is a self-supervised vision-language pretraining method that significantly improves medical VQA performance by learning multimodal representations through masked modeling and contrastive learning. However, its ability to predict free-form answers may be influenced by the complexity of the dataset. 
BiomedCLIP~\cite{zhang2024biomedclipmultimodalbiomedicalfoundation} is a multimodal biomedical foundation model trained on PMC-15M. It outperforms earlier vision-language models like PubMedCLIP~\cite{eslami-etal-2023-pubmedclip} and MedCLIP, as well as radiology-specific models such as BioViL~\cite{Boecking_2022}.
Bazi ~\cite{bazi_vqa_2023_1} proposes a vision-language model based on a Transformer encoder-decoder architecture. It leverages the CLIP model for semantic embedding and uses a generative decoder to produce answers in an autoregressive manner. 
Recently, inspired by LLaVA \citep{liu2024visual}, 
LLaVA-Med \citep{li2024llava} fine-tunes LLaVA by self-generated biomedical instruction-following dataset to address challenges in medical image interpretation. However, the explicit handling of region-specific information in complex medical scenarios remains insufficiently explored.

\section{Conclusion}
In this paper, we introduce R-LLaVA, an effective method to enhance Med-VQA understanding by leveraging Visual Regions of Interest. This approach integrates simple physician annotations, such as bounding boxes, as prior knowledge, directly infusing these visual cues into the image space via CLIP. During training, these annotated Visual Regions of Interest are fed into the LLaVA model to boost Biomedical VQA Understanding. A specially constructed multiple-choice dataset demonstrates the positive impact of Visual Regions of Interest within R-LLaVA. Experimental results on four Med-VQA datasets show that R-LLaVA outperforms existing SoTA techniques, significantly surpassing recent methods.

\section*{Limitation} 
Based on our experiments, we demonstrate that even minimal annotations provided by doctors can significantly enhance the accuracy of \ourmodel{}. While doctor-specified regions of interest are not required during inference—\ourmodel{} is capable of processing both annotated and non-annotated cases—some degree of annotation is necessary during the training phase to achieve optimal performance. 
Additionally, we conducted a series of ablation studies examining model configurations, training strategies, and the inclusion of visual regions of interest. These studies consistently showed that the proposed method outperforms baseline approaches, particularly in clinical scenarios where visual region annotations are utilized.

\section*{Potential Risks and Broader Impacts}

The proposed method, which builds on LLaVA \citep{liu2024visual,cai2024vip}, inherits several inherent challenges associated with VLLMs, such as hallucination and biases. 
Regarding hallucination, as demonstrated in \cref{tab:main_table} and \cref{fig:result_example}, \ourmodel{} makes more effective use of the visual regions of interest, leading to generally superior performance. 
Nevertheless, it may still produce responses that are not grounded in factual information or the input data. We consider this work a significant step toward enhancing biomedical VQA by utilizing doctor guidance. 
To address potential biases \citep{ruan2023causal}, particularly those that could favor or disfavor specific demographics, we test \ourmodel{} on cases where annotations were unavailable due to privacy constraints. Additionally, integrating improved vision or language encoders could further alleviate these biases. 
Given the high computational cost and energy demand of training LLMs in general, we leverage pre-trained image and language encoders. This approach is especially appropriate given the relatively small size of biomedical datasets, eliminating the need for training from scratch.

In particular, \ourmodel{} could democratize access to expert-level biomedical information, offering non-specialists—such as primary care physicians, medical trainees, and even patients themselves—an enhanced tool for understanding complex medical data. This capability is crucial in low-resource settings where access to specialized medical expertise is limited, thus reducing healthcare disparities on a global scale. Furthermore, by enabling more accurate and contextually grounded responses, the model could assist clinicians in making more informed, evidence-based decisions, potentially improving patient outcomes and reducing diagnostic errors. 

In summary, the proposed method not only advances the technical status of biomedical VQA but also holds promise for generating significant social benefits, particularly by improving healthcare accessibility, reducing disparities, and promoting AI development. These contributions align with the growing emphasis on the ethical and socially responsible deployment of AI in sensitive domains such as healthcare.

\newpage

\bibliography{iclr2025_conference}

\begin{thebibliography}{55}
\providecommand{\natexlab}[1]{#1}
\providecommand{\url}[1]{\texttt{#1}}
\expandafter\ifx\csname urlstyle\endcsname\relax
  \providecommand{\doi}[1]{doi: #1}\else
  \providecommand{\doi}{doi: \begingroup \urlstyle{rm}\Url}\fi

\bibitem[Alayrac et~al.(2022)Alayrac, Donahue, Luc, Miech, Barr, Hasson, Lenc, Mensch, Millican, Reynolds, et~al.]{alayrac2022flamingo}
Jean-Baptiste Alayrac, Jeff Donahue, Pauline Luc, Antoine Miech, Iain Barr, Yana Hasson, Karel Lenc, Arthur Mensch, Katherine Millican, Malcolm Reynolds, et~al.
\newblock Flamingo: a visual language model for few-shot learning.
\newblock \emph{Advances in neural information processing systems}, 35:\penalty0 23716--23736, 2022.

\bibitem[Banerjee et~al.(2021)Banerjee, Gokhale, Yang, and Baral]{banerjee2020weaqa}
Pratyay Banerjee, Tejas Gokhale, Yezhou Yang, and Chitta Baral.
\newblock Weaqa: Weak supervision via captions for visual question answering.
\newblock \emph{Findings of the Association for Computational Linguistics: ACL-IJCNLP 2021}, 2021.

\bibitem[Bazi et~al.(2023{\natexlab{a}})Bazi, Al~Rahhal, Bashmal, and Zuair]{bazi_vqa_2023_1}
Yakoub Bazi, Mohamad Al~Rahhal, Laila Bashmal, and Mansour Zuair.
\newblock Vision–language model for visual question answering in medical imagery.
\newblock \emph{Bioengineering}, 10, 03 2023{\natexlab{a}}.
\newblock \doi{10.3390/bioengineering10030380}.

\bibitem[Bazi et~al.(2023{\natexlab{b}})Bazi, Rahhal, Bashmal, and Zuair]{bazi2023vision}
Yakoub Bazi, Mohamad Mahmoud~Al Rahhal, Laila Bashmal, and Mansour Zuair.
\newblock Vision--language model for visual question answering in medical imagery.
\newblock \emph{Bioengineering}, 10\penalty0 (3):\penalty0 380, 2023{\natexlab{b}}.

\bibitem[Boecking et~al.(2022)Boecking, Usuyama, Bannur, Castro, Schwaighofer, Hyland, Wetscherek, Naumann, Nori, Alvarez-Valle, Poon, and Oktay]{Boecking_2022}
Benedikt Boecking, Naoto Usuyama, Shruthi Bannur, Daniel~C. Castro, Anton Schwaighofer, Stephanie Hyland, Maria Wetscherek, Tristan Naumann, Aditya Nori, Javier Alvarez-Valle, Hoifung Poon, and Ozan Oktay.
\newblock \emph{Making the Most of Text Semantics to Improve Biomedical Vision–Language Processing}, pp.\  1–21.
\newblock Springer Nature Switzerland, 2022.
\newblock ISBN 9783031200595.
\newblock \doi{10.1007/978-3-031-20059-5\_1}.
\newblock URL \url{http://dx.doi.org/10.1007/978-3-031-20059-5_1}.

\bibitem[Brown et~al.(2020)Brown, Mann, Ryder, Subbiah, Kaplan, Dhariwal, Neelakantan, Shyam, Sastry, Askell, et~al.]{brown2020language}
Tom Brown, Benjamin Mann, Nick Ryder, Melanie Subbiah, Jared~D Kaplan, Prafulla Dhariwal, Arvind Neelakantan, Pranav Shyam, Girish Sastry, Amanda Askell, et~al.
\newblock Language models are few-shot learners.
\newblock \emph{Advances in neural information processing systems}, 33:\penalty0 1877--1901, 2020.

\bibitem[Cai et~al.(2024)Cai, Liu, Mustikovela, Meyer, Chai, Park, and Lee]{cai2024vip}
Mu~Cai, Haotian Liu, Siva~Karthik Mustikovela, Gregory~P Meyer, Yuning Chai, Dennis Park, and Yong~Jae Lee.
\newblock Vip-llava: Making large multimodal models understand arbitrary visual prompts.
\newblock In \emph{Proceedings of the IEEE/CVF Conference on Computer Vision and Pattern Recognition}, pp.\  12914--12923, 2024.

\bibitem[Changpinyo et~al.(2022)Changpinyo, Kukliansy, Szpektor, Chen, Ding, and Soricut]{changpinyo2022all}
Soravit Changpinyo, Doron Kukliansy, Idan Szpektor, Xi~Chen, Nan Ding, and Radu Soricut.
\newblock All you may need for vqa are image captions.
\newblock In \emph{Proceedings of the 2022 Conference of the North American Chapter of the Association for Computational Linguistics: Human Language Technologies}, pp.\  1947--1963, 2022.

\bibitem[Chen et~al.(2024)Chen, Jie, and Ma]{chen2024llava}
Shaoxiang Chen, Zequn Jie, and Lin Ma.
\newblock Llava-mole: Sparse mixture of lora experts for mitigating data conflicts in instruction finetuning mllms.
\newblock \emph{arXiv preprint arXiv:2401.16160}, 2024.

\bibitem[Chen et~al.(2022{\natexlab{a}})Chen, Wang, Changpinyo, Piergiovanni, Padlewski, Salz, Goodman, Grycner, Mustafa, Beyer, et~al.]{chen2022pali}
Xi~Chen, Xiao Wang, Soravit Changpinyo, AJ~Piergiovanni, Piotr Padlewski, Daniel Salz, Sebastian Goodman, Adam Grycner, Basil Mustafa, Lucas Beyer, et~al.
\newblock Pali: A jointly-scaled multilingual language-image model.
\newblock \emph{arXiv preprint arXiv:2209.06794}, 2022{\natexlab{a}}.

\bibitem[Chen et~al.(2022{\natexlab{b}})Chen, Li, and Wan]{chen2022align}
Zhihong Chen, Guanbin Li, and Xiang Wan.
\newblock Align, reason and learn: Enhancing medical vision-and-language pre-training with knowledge.
\newblock In \emph{Proceedings of the 30th ACM International Conference on Multimedia}, pp.\  5152--5161, 2022{\natexlab{b}}.

\bibitem[Chiang et~al.(2023)Chiang, Li, Lin, Sheng, Wu, Zhang, Zheng, Zhuang, Zhuang, Gonzalez, et~al.]{chiang2023vicuna}
Wei-Lin Chiang, Zhuohan Li, Zi~Lin, Ying Sheng, Zhanghao Wu, Hao Zhang, Lianmin Zheng, Siyuan Zhuang, Yonghao Zhuang, Joseph~E Gonzalez, et~al.
\newblock Vicuna: An open-source chatbot impressing gpt-4 with 90\%* chatgpt quality.
\newblock \emph{See https://vicuna. lmsys. org (accessed 14 April 2023)}, 2\penalty0 (3):\penalty0 6, 2023.

\bibitem[Dosovitskiy et~al.(2021)Dosovitskiy, Beyer, Kolesnikov, Weissenborn, Zhai, Unterthiner, Dehghani, Minderer, Heigold, Gelly, Uszkoreit, and Houlsby]{dosovitskiy2020vit}
Alexey Dosovitskiy, Lucas Beyer, Alexander Kolesnikov, Dirk Weissenborn, Xiaohua Zhai, Thomas Unterthiner, Mostafa Dehghani, Matthias Minderer, Georg Heigold, Sylvain Gelly, Jakob Uszkoreit, and Neil Houlsby.
\newblock An image is worth 16x16 words: Transformers for image recognition at scale.
\newblock \emph{ICLR}, 2021.

\bibitem[Eslami et~al.(2021)Eslami, de~Melo, and Meinel]{eslami2021does}
Sedigheh Eslami, Gerard de~Melo, and Christoph Meinel.
\newblock Does clip benefit visual question answering in the medical domain as much as it does in the general domain?
\newblock \emph{arXiv preprint arXiv:2112.13906}, 2021.

\bibitem[Eslami et~al.(2023{\natexlab{a}})Eslami, Meinel, and de~Melo]{eslami-etal-2023-pubmedclip}
Sedigheh Eslami, Christoph Meinel, and Gerard de~Melo.
\newblock {P}ub{M}ed{CLIP}: How much does {CLIP} benefit visual question answering in the medical domain?
\newblock In Andreas Vlachos and Isabelle Augenstein (eds.), \emph{Findings of the Association for Computational Linguistics: EACL 2023}, pp.\  1181--1193, Dubrovnik, Croatia, May 2023{\natexlab{a}}. Association for Computational Linguistics.
\newblock \doi{10.18653/v1/2023.findings-eacl.88}.
\newblock URL \url{https://aclanthology.org/2023.findings-eacl.88}.

\bibitem[Eslami et~al.(2023{\natexlab{b}})Eslami, Meinel, and De~Melo]{eslami2023pubmedclip}
Sedigheh Eslami, Christoph Meinel, and Gerard De~Melo.
\newblock Pubmedclip: How much does clip benefit visual question answering in the medical domain?
\newblock In \emph{Findings of the Association for Computational Linguistics: EACL 2023}, pp.\  1151--1163, 2023{\natexlab{b}}.

\bibitem[Gai et~al.(2024)Gai, Zhou, Liu, Feng, Wu, and Liu]{gai2024medthink}
Xiaotang Gai, Chenyi Zhou, Jiaxiang Liu, Yang Feng, Jian Wu, and Zuozhu Liu.
\newblock Medthink: Explaining medical visual question answering via multimodal decision-making rationale.
\newblock \emph{arXiv preprint arXiv:2404.12372}, 2024.

\bibitem[Gong et~al.(2021)Gong, Chen, Liu, Yu, and Li]{gong2021cross}
Haifan Gong, Guanqi Chen, Sishuo Liu, Yizhou Yu, and Guanbin Li.
\newblock Cross-modal self-attention with multi-task pre-training for medical visual question answering.
\newblock In \emph{Proceedings of the 2021 International Conference on Multimedia Retrieval}, pp.\  456--460, 2021.

\bibitem[Jiang et~al.(2024)Jiang, Sablayrolles, Roux, Mensch, Savary, Bamford, Chaplot, Casas, Hanna, Bressand, et~al.]{jiang2024mixtral}
Albert~Q Jiang, Alexandre Sablayrolles, Antoine Roux, Arthur Mensch, Blanche Savary, Chris Bamford, Devendra~Singh Chaplot, Diego de~las Casas, Emma~Bou Hanna, Florian Bressand, et~al.
\newblock Mixtral of experts.
\newblock \emph{arXiv preprint arXiv:2401.04088}, 2024.

\bibitem[Khare et~al.(2021)Khare, Bagal, Mathew, Devi, Priyakumar, and Jawahar]{khare2021mmbert}
Yash Khare, Viraj Bagal, Minesh Mathew, Adithi Devi, U~Deva Priyakumar, and CV~Jawahar.
\newblock Mmbert: Multimodal bert pretraining for improved medical vqa.
\newblock In \emph{2021 IEEE 18th International Symposium on Biomedical Imaging (ISBI)}, pp.\  1033--1036. IEEE, 2021.

\bibitem[Kim et~al.(2018)Kim, Jun, and Zhang]{kim2018bilinear}
Jin-Hwa Kim, Jaehyun Jun, and Byoung-Tak Zhang.
\newblock Bilinear attention networks.
\newblock \emph{Advances in neural information processing systems}, 31, 2018.

\bibitem[Li et~al.(2023)Li, Wong, Zhang, Usuyama, Liu, Yang, Naumann, Poon, and Gao]{li2023llavamedtraininglargelanguageandvision}
Chunyuan Li, Cliff Wong, Sheng Zhang, Naoto Usuyama, Haotian Liu, Jianwei Yang, Tristan Naumann, Hoifung Poon, and Jianfeng Gao.
\newblock Llava-med: Training a large language-and-vision assistant for biomedicine in one day, 2023.
\newblock URL \url{https://arxiv.org/abs/2306.00890}.

\bibitem[Li et~al.(2024)Li, Wong, Zhang, Usuyama, Liu, Yang, Naumann, Poon, and Gao]{li2024llava}
Chunyuan Li, Cliff Wong, Sheng Zhang, Naoto Usuyama, Haotian Liu, Jianwei Yang, Tristan Naumann, Hoifung Poon, and Jianfeng Gao.
\newblock Llava-med: Training a large language-and-vision assistant for biomedicine in one day.
\newblock \emph{Advances in Neural Information Processing Systems}, 36, 2024.

\bibitem[Li et~al.(2019)Li, Yatskar, Yin, Hsieh, and Chang]{li2019visualbert}
Liunian~Harold Li, Mark Yatskar, Da~Yin, Cho-Jui Hsieh, and Kai-Wei Chang.
\newblock Visualbert: A simple and performant baseline for vision and language.
\newblock \emph{arXiv preprint arXiv:1908.03557}, 2019.

\bibitem[Li et~al.(2022{\natexlab{a}})Li, Liu, Tan, Liao, and Zhong]{li2022self}
Pengfei Li, Gang Liu, Lin Tan, Jinying Liao, and Shenjun Zhong.
\newblock Self-supervised vision-language pretraining for medical visual question answering.
\newblock \emph{arXiv preprint arXiv:2211.13594}, 2022{\natexlab{a}}.

\bibitem[Li et~al.(2022{\natexlab{b}})Li, Liu, Tan, Liao, and Zhong]{li2022selfsupervisedvisionlanguagepretrainingmedical}
Pengfei Li, Gang Liu, Lin Tan, Jinying Liao, and Shenjun Zhong.
\newblock Self-supervised vision-language pretraining for medical visual question answering, 2022{\natexlab{b}}.
\newblock URL \url{https://arxiv.org/abs/2211.13594}.

\bibitem[Liu et~al.(2024{\natexlab{a}})Liu, Li, Wu, and Lee]{liu2024visual}
Haotian Liu, Chunyuan Li, Qingyang Wu, and Yong~Jae Lee.
\newblock Visual instruction tuning.
\newblock \emph{Advances in neural information processing systems}, 36, 2024{\natexlab{a}}.

\bibitem[Liu et~al.(2023{\natexlab{a}})Liu, Hu, Zhang, Feng, Hao, Lv, and Liu]{liu2023parameter}
Jiaxiang Liu, Tianxiang Hu, Yan Zhang, Yang Feng, Jin Hao, Junhui Lv, and Zuozhu Liu.
\newblock Parameter-efficient transfer learning for medical visual question answering.
\newblock \emph{IEEE Transactions on Emerging Topics in Computational Intelligence}, 2023{\natexlab{a}}.

\bibitem[Liu et~al.(2023{\natexlab{b}})Liu, Hu, Zhang, Gai, FENG, and Liu]{liu2023chatgpt}
Jiaxiang Liu, Tianxiang Hu, Yan Zhang, Xiaotang Gai, YANG FENG, and Zuozhu Liu.
\newblock A chatgpt aided explainable framework for zero-shot medical image diagnosis.
\newblock In \emph{ICML 3rd Workshop on Interpretable Machine Learning in Healthcare (IMLH)}, 2023{\natexlab{b}}.

\bibitem[Liu et~al.(2024{\natexlab{b}})Liu, Hu, Xiong, Du, Feng, Wu, Zhou, and Liu]{liu2024vpl}
Jiaxiang Liu, Tianxiang Hu, Huimin Xiong, Jiawei Du, Yang Feng, Jian Wu, Joey Zhou, and Zuozhu Liu.
\newblock Vpl: Visual proxy learning framework for zero-shot medical image diagnosis.
\newblock In \emph{Findings of the Association for Computational Linguistics: EMNLP 2024}, pp.\  9978--9992, 2024{\natexlab{b}}.

\bibitem[Liu et~al.(2024{\natexlab{c}})Liu, Wang, Du, Zhou, and Liu]{liu-etal-2024-medcot}
Jiaxiang Liu, Yuan Wang, Jiawei Du, Joey~Tianyi Zhou, and Zuozhu Liu.
\newblock {M}ed{C}o{T}: Medical chain of thought via hierarchical expert.
\newblock In Yaser Al-Onaizan, Mohit Bansal, and Yun-Nung Chen (eds.), \emph{Proceedings of the 2024 Conference on Empirical Methods in Natural Language Processing}, pp.\  17371--17389, Miami, Florida, USA, November 2024{\natexlab{c}}. Association for Computational Linguistics.
\newblock \doi{10.18653/v1/2024.emnlp-main.962}.
\newblock URL \url{https://aclanthology.org/2024.emnlp-main.962/}.

\bibitem[Liu et~al.(2025)Liu, Hu, Du, Zhang, Zhou, and Liu]{liu2025kpl}
Jiaxiang Liu, Tianxiang Hu, Jiawei Du, Ruiyuan Zhang, Joey~Tianyi Zhou, and Zuozhu Liu.
\newblock Kpl: Training-free medical knowledge mining of vision-language models.
\newblock \emph{arXiv preprint arXiv:2501.11231}, 2025.

\bibitem[Liu et~al.(2023{\natexlab{c}})Liu, Wang, Xu, and Zhou]{liu2023q2atransformer}
Yunyi Liu, Zhanyu Wang, Dong Xu, and Luping Zhou.
\newblock Q2atransformer: Improving medical vqa via an answer querying decoder.
\newblock \emph{arXiv preprint arXiv:2304.01611}, 2023{\natexlab{c}}.

\bibitem[Lu et~al.(2019)Lu, Batra, Parikh, and Lee]{lu2019vilbert}
Jiasen Lu, Dhruv Batra, Devi Parikh, and Stefan Lee.
\newblock Vilbert: Pretraining task-agnostic visiolinguistic representations for vision-and-language tasks.
\newblock \emph{Advances in neural information processing systems}, 32, 2019.

\bibitem[Nguyen et~al.(2019)Nguyen, Do, Nguyen, Do, Tjiputra, and Tran]{nguyen2019overcoming}
Binh~D Nguyen, Thanh-Toan Do, Binh~X Nguyen, Tuong Do, Erman Tjiputra, and Quang~D Tran.
\newblock Overcoming data limitation in medical visual question answering.
\newblock In \emph{International Conference on Medical Image Computing and Computer-Assisted Intervention}, pp.\  522--530. Springer, 2019.

\bibitem[Pelka et~al.(2018)Pelka, Koitka, R{\"u}ckert, Nensa, and Friedrich]{pelka2018radiology}
Obioma Pelka, Sven Koitka, Johannes R{\"u}ckert, Felix Nensa, and Christoph~M Friedrich.
\newblock Radiology objects in context (roco): a multimodal image dataset.
\newblock In \emph{Intravascular Imaging and Computer Assisted Stenting and Large-Scale Annotation of Biomedical Data and Expert Label Synthesis: 7th Joint International Workshop, CVII-STENT 2018 and Third International Workshop, LABELS 2018, Held in Conjunction with MICCAI 2018, Granada, Spain, September 16, 2018, Proceedings 3}, pp.\  180--189. Springer, 2018.

\bibitem[Radford et~al.(2021)Radford, Kim, Hallacy, Ramesh, Goh, Agarwal, Sastry, Askell, Mishkin, Clark, et~al.]{radford2021learning}
Alec Radford, Jong~Wook Kim, Chris Hallacy, Aditya Ramesh, Gabriel Goh, Sandhini Agarwal, Girish Sastry, Amanda Askell, Pamela Mishkin, Jack Clark, et~al.
\newblock Learning transferable visual models from natural language supervision.
\newblock In \emph{International conference on machine learning}, pp.\  8748--8763. PMLR, 2021.

\bibitem[Ren \& Zhou(2020)Ren and Zhou]{ren2020cgmvqa}
Fuji Ren and Yangyang Zhou.
\newblock Cgmvqa: A new classification and generative model for medical visual question answering.
\newblock \emph{IEEE Access}, 8:\penalty0 50626--50636, 2020.

\bibitem[Ruan \& Di(2024)Ruan and Di]{ruan2024infostgcan}
Kangrui Ruan and Xuan Di.
\newblock Infostgcan: An information-maximizing spatial-temporal graph convolutional attention network for heterogeneous human trajectory prediction.
\newblock \emph{Computers}, 13\penalty0 (6):\penalty0 151, 2024.

\bibitem[Ruan et~al.(2023)Ruan, Zhang, Di, and Bareinboim]{ruan2023causal}
Kangrui Ruan, Junzhe Zhang, Xuan Di, and Elias Bareinboim.
\newblock Causal imitation learning via inverse reinforcement learning.
\newblock In \emph{The Eleventh International Conference on Learning Representations}, 2023.

\bibitem[Ruan et~al.(2024{\natexlab{a}})Ruan, He, Wang, Zhou, Feng, and Kebarighotbi]{ruan2024s2e}
Kangrui Ruan, Xin He, Jiyang Wang, Xiaozhou Zhou, Helian Feng, and Ali Kebarighotbi.
\newblock S2e: Towards an end-to-end entity resolution solution from acoustic signal.
\newblock In \emph{ICASSP 2024-2024 IEEE International Conference on Acoustics, Speech and Signal Processing (ICASSP)}, pp.\  10441--10445. IEEE, 2024{\natexlab{a}}.

\bibitem[Ruan et~al.(2024{\natexlab{b}})Ruan, Wang, and Di]{ruan2024twitter}
Kangrui Ruan, Xinyang Wang, and Xuan Di.
\newblock From twitter to reasoner: Understand mobility travel modes and sentiment using large language models.
\newblock \emph{arXiv preprint arXiv:2411.02666}, 2024{\natexlab{b}}.

\bibitem[Ruan et~al.(2025)Ruan, Zhang, Di, and Bareinboim]{ruan2025causal}
Kangrui Ruan, Junzhe Zhang, Xuan Di, and Elias Bareinboim.
\newblock Causal imitation for markov decision processes: A partial identification approach.
\newblock \emph{Advances in Neural Information Processing Systems}, 37:\penalty0 87592--87620, 2025.

\bibitem[Song et~al.(2022)Song, Dong, Zhang, Liu, and Wei]{song2022clip}
Haoyu Song, Li~Dong, Weinan Zhang, Ting Liu, and Furu Wei.
\newblock Clip models are few-shot learners: Empirical studies on vqa and visual entailment.
\newblock In \emph{Proceedings of the 60th Annual Meeting of the Association for Computational Linguistics (Volume 1: Long Papers)}, pp.\  6088--6100, 2022.

\bibitem[Team et~al.(2023)Team, Anil, Borgeaud, Wu, Alayrac, Yu, Soricut, Schalkwyk, Dai, Hauth, et~al.]{team2023gemini}
Gemini Team, Rohan Anil, Sebastian Borgeaud, Yonghui Wu, Jean-Baptiste Alayrac, Jiahui Yu, Radu Soricut, Johan Schalkwyk, Andrew~M Dai, Anja Hauth, et~al.
\newblock Gemini: a family of highly capable multimodal models.
\newblock \emph{arXiv preprint arXiv:2312.11805}, 2023.

\bibitem[Tiong et~al.(2022)Tiong, Li, Li, Savarese, and Hoi]{tiong2022plug}
Anthony Meng~Huat Tiong, Junnan Li, Boyang Li, Silvio Savarese, and Steven~C.H. Hoi.
\newblock Plug-and-play {VQA}: Zero-shot {VQA} by conjoining large pretrained models with zero training.
\newblock In \emph{Findings of the Association for Computational Linguistics: EMNLP 2022}, pp.\  951--967, Abu Dhabi, United Arab Emirates, December 2022. Association for Computational Linguistics.
\newblock URL \url{https://aclanthology.org/2022.findings-emnlp.67}.

\bibitem[Van~Sonsbeek et~al.(2023)Van~Sonsbeek, Derakhshani, Najdenkoska, Snoek, and Worring]{van2023open}
Tom Van~Sonsbeek, Mohammad~Mahdi Derakhshani, Ivona Najdenkoska, Cees~GM Snoek, and Marcel Worring.
\newblock Open-ended medical visual question answering through prefix tuning of language models.
\newblock In \emph{International Conference on Medical Image Computing and Computer-Assisted Intervention}, pp.\  726--736. Springer, 2023.

\bibitem[Vaswani et~al.(2017)Vaswani, Shazeer, Parmar, Uszkoreit, Jones, Gomez, Kaiser, and Polosukhin]{vaswani2017attention}
Ashish Vaswani, Noam Shazeer, Niki Parmar, Jakob Uszkoreit, Llion Jones, Aidan~N Gomez, {\L}ukasz Kaiser, and Illia Polosukhin.
\newblock Attention is all you need.
\newblock \emph{Advances in neural information processing systems}, 30, 2017.

\bibitem[Wang et~al.(2022)Wang, Xiao, Codella, Yang, Chen, Zhou, Chang, Dai, You, and Yuan]{wang2022clip}
Zhecan Wang, Bin Xiao, Noel Codella, Jianwei Yang, Yen-Chun Chen, Luowei Zhou, Shih-Fu Chang, Xiyang Dai, Haoxuan You, and Lu~Yuan.
\newblock Clip-td: Clip targeted distillation for vision-language tasks.
\newblock In \emph{International Conference on Learning Representations}, 2022.

\bibitem[Yang et~al.(2024)Yang, Xu, Sellergren, Kohlberger, Zhou, Ktena, Kiraly, Ahmed, Hormozdiari, Jaroensri, et~al.]{yang2024advancing}
Lin Yang, Shawn Xu, Andrew Sellergren, Timo Kohlberger, Yuchen Zhou, Ira Ktena, Atilla Kiraly, Faruk Ahmed, Farhad Hormozdiari, Tiam Jaroensri, et~al.
\newblock Advancing multimodal medical capabilities of gemini.
\newblock \emph{arXiv preprint arXiv:2405.03162}, 2024.

\bibitem[Zhan et~al.(2020)Zhan, Liu, Fan, Chen, and Wu]{zhan2020medical}
Li-Ming Zhan, Bo~Liu, Lu~Fan, Jiaxin Chen, and Xiao-Ming Wu.
\newblock Medical visual question answering via conditional reasoning.
\newblock In \emph{Proceedings of the 28th ACM International Conference on Multimedia}, pp.\  2345--2354, 2020.

\bibitem[Zhang et~al.(2023{\natexlab{a}})Zhang, Yu, Yan, Liu, Adhikarla, Fu, Chen, Chen, Zhou, Li, et~al.]{zhang2023biomedgpt}
Kai Zhang, Jun Yu, Zhiling Yan, Yixin Liu, Eashan Adhikarla, Sunyang Fu, Xun Chen, Chen Chen, Yuyin Zhou, Xiang Li, et~al.
\newblock Biomedgpt: A unified and generalist biomedical generative pre-trained transformer for vision, language, and multimodal tasks.
\newblock \emph{arXiv preprint arXiv:2305.17100}, 2023{\natexlab{a}}.

\bibitem[Zhang et~al.(2023{\natexlab{b}})Zhang, Xu, Usuyama, Bagga, Tinn, Preston, Rao, Wei, Valluri, Wong, et~al.]{zhang2023large}
Sheng Zhang, Yanbo Xu, Naoto Usuyama, Jaspreet Bagga, Robert Tinn, Sam Preston, Rajesh Rao, Mu~Wei, Naveen Valluri, Cliff Wong, et~al.
\newblock Large-scale domain-specific pretraining for biomedical vision-language processing.
\newblock \emph{arXiv preprint arXiv:2303.00915}, 2023{\natexlab{b}}.

\bibitem[Zhang et~al.(2024)Zhang, Xu, Usuyama, Xu, Bagga, Tinn, Preston, Rao, Wei, Valluri, Wong, Tupini, Wang, Mazzola, Shukla, Liden, Gao, Lungren, Naumann, Wang, and Poon]{zhang2024biomedclipmultimodalbiomedicalfoundation}
Sheng Zhang, Yanbo Xu, Naoto Usuyama, Hanwen Xu, Jaspreet Bagga, Robert Tinn, Sam Preston, Rajesh Rao, Mu~Wei, Naveen Valluri, Cliff Wong, Andrea Tupini, Yu~Wang, Matt Mazzola, Swadheen Shukla, Lars Liden, Jianfeng Gao, Matthew~P. Lungren, Tristan Naumann, Sheng Wang, and Hoifung Poon.
\newblock Biomedclip: a multimodal biomedical foundation model pretrained from fifteen million scientific image-text pairs, 2024.
\newblock URL \url{https://arxiv.org/abs/2303.00915}.

\bibitem[Zheng et~al.(2023)Zheng, Chiang, Sheng, Zhuang, Wu, Zhuang, Lin, Li, Li, Xing, Zhang, Gonzalez, and Stoica]{zheng2023judging}
Lianmin Zheng, Wei-Lin Chiang, Ying Sheng, Siyuan Zhuang, Zhanghao Wu, Yonghao Zhuang, Zi~Lin, Zhuohan Li, Dacheng Li, Eric.~P Xing, Hao Zhang, Joseph~E. Gonzalez, and Ion Stoica.
\newblock Judging llm-as-a-judge with mt-bench and chatbot arena, 2023.

\end{thebibliography}
\bibliographystyle{iclr2025_conference}





\end{document}